\pdfoutput=1

\documentclass[11pt]{article}

\usepackage[final]{acl}

\usepackage{amsmath}
\usepackage{amssymb}
\usepackage{times}
\usepackage{latexsym}
\usepackage{caption}
\usepackage{subcaption}
\usepackage{graphicx}
\usepackage{booktabs} 

\usepackage[T1]{fontenc}

\usepackage[utf8]{inputenc}

\usepackage{microtype}

\usepackage{inconsolata}

\usepackage{graphicx}

\title{Novel-WD: Exploring acquisition of Novel World Knowledge in LLMs Using Prefix-Tuning}

\author{Maxime Méloux \and Christophe Cerisara \\
Université de Lorraine, CNRS, LORIA, Nancy, France}

\begin{document}

\maketitle

\begin{abstract}
Teaching new information to pre-trained large language models (PLM) is a crucial but challenging task. Model adaptation techniques, such as fine-tuning and parameter-efficient training have been
shown to store new facts at a slow rate; continual learning is an option but is costly and prone to catastrophic forgetting.
This work studies and quantifies how PLM may learn and remember new world knowledge facts that do not occur in their pre-training corpus, which only contains world knowledge up to a certain date.
To that purpose, we first propose \textsc{Novel-WD}, a new dataset consisting of sentences containing novel facts extracted from recent Wikidata updates, along with two evaluation tasks in the form of causal language modeling and multiple choice questions (MCQ). We make this dataset freely available to the community, and release a procedure to later build new versions of similar datasets with up-to-date information.
We also explore the use of prefix-tuning for novel information learning, and analyze how much information can be stored within a given prefix. We show that a single fact can reliably be encoded within a single prefix, and that the prefix capacity increases with its length and with the base model size.
\end{abstract}

\section{Introduction}
Pre-trained language models (PLM or LLM)~\citep{chiang_recent_2022} are typically trained on raw texts with a self-supervised loss and further
adapted to downstream tasks with, e.g., finetuning~\citep{dai_semi-supervised_2015,howard_universal_2018,radford_language_2019}.
Hence, the world knowledge that PLM have acquired is prior to the cut-off date of their pretraining corpus~\citep{alivanistos_prompting_2022,kucharavy_fundamentals_2023}. 
A major challenge is then how to reliably teach PLMs novel factual knowledge.
Fine-tuning has been one of the main proposed approaches to adapt pre-trained models to new tasks and domains. However, full model fine-tuning can lead to catastrophic forgetting~\citep{french_catastrophic_1999,kirkpatrick_overcoming_2017}, and can be costly when performed on large models~\citep{strubell_energy_2020}. Furthermore, \citet{wei_simple_2023} showed that when fine-tuning a model on a small corpus with new information, the model may instead learn to hallucinate unseen facts.
Parameter-efficient fine-tuning (PEFT) methods have emerged as an lightweight alternative to full model fine-tuning, in which only a fraction of the parameters of the original model are modified. PEFT allows for efficiently modifying a small fraction of model parameters using methods such as prefix-tuning~\citep{li_prefix-tuning_2021}, adapter-tuning~\citep{he_towards_2021} or LoRA~\citep{hu_lora_2021}. In-context learning~\citep{logan_iv_cutting_2022}, prompting~\citep{liu_pre-train_2023} and prompt-tuning~\citep{lester_power_2021} are currently amongst the most reliable ways to inject new knowledge in PLM.

In this study, we focus on prefix-tuning~\citep{li_prefix-tuning_2021}, a fine-tuning method in which the pre-trained model parameters are kept frozen, but a few small continuous vectors called the \textit{prefix} are optimized. Based on the idea that context can steer a language model without changing its parameters, prefix-tuning optimizes the model's context as one or several continuous vectors corresponding to either embeddings or to key-query pairs in attention layers, whose effects will be propagated to all activation layers and subsequent tokens.

\citet{wang_knowledge_2022} and \citet{liu_generated_2022} showed that novel knowledge can efficiently be contextually fed into large language models through prompting. However, the size of a prompt in a given model is limited by the context size of that model. In this paper, we view prefix-tuning as a generalized form of prompting taking continuous values, and having controllable depth and length, and as such, we hypothesize that this method can reliably store significant amounts of factual information. This is backed by the findings of \citet{kossen_-context_2023}, which argue that in-context learning enables a model to learn information. Our goal is therefore to investigate this question in the case of prefix-tuning, and more specifically how much knowledge can be compressed into the prefix. In addition, by using prefix-tuning rather than LoRA, fine-tuning or adapters, we hope to avoid the hallucination problem mentioned in \cite{wei_simple_2023} by working with (generalized) prompts without modifying the existing model weights.

\begin{figure*}
    \centering
    \includegraphics[width=0.8\linewidth]{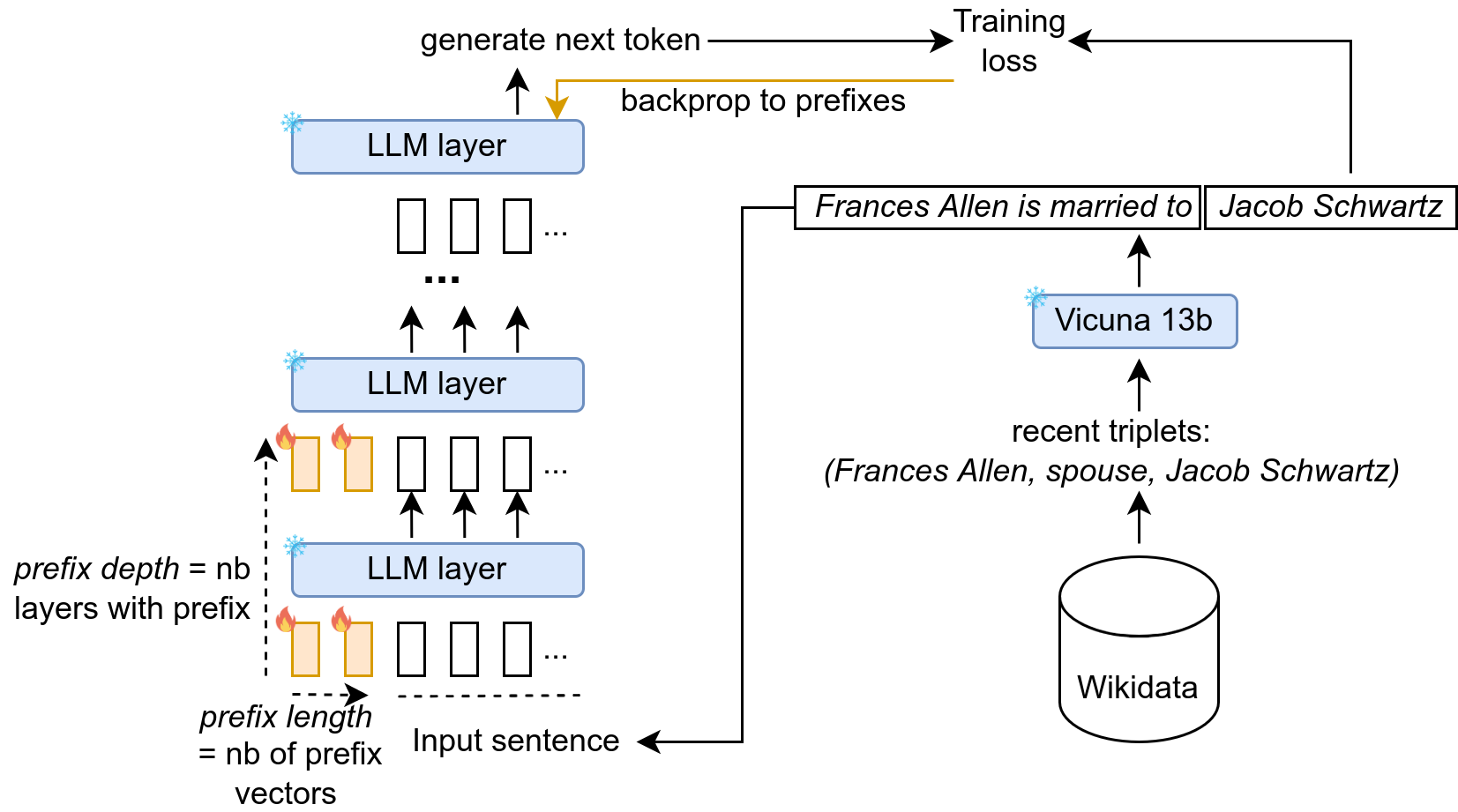}
    \caption{Proposed approach: new facts are extracted from Wikidata, transformed into sentences with Vicuna-13b and trained into prefixes. We claim and show that this architecture is better than LoRA to capture novel knowledge.}
    \label{fig1}
\end{figure*}

Figure~\ref{fig1} summarizes our proposed approach, which exploits recent Wikidata updates to automatically generate a corpus of new facts: \textsc{Novel-WD}. We then propose a nearly automatic procedure to create a dynamic benchmark from
this corpus of facts that evaluates updated LLMs in terms of perplexity, new facts generation and accuracy on multiple-choice question-answering. We then
evaluate and show that prefix-tuning performs better than LoRA for new facts learning on this dataset.

\section{Related work}
Adapting models to new tasks is a relatively old problem. \citet{yoon_lifelong_2018} showed that dynamically expandable networks can obtain good performance in this setting by slowly increasing model capacity. \citet{lin_continual_2022} explored the task of improving accuracy of Transformer models on out-of-data streams using continual model refinement (CMR) to maximize the diversity of training samples in a non-stationary distribution. \citet{razdaibiedina_progressive_2023} showed that using a collection of progressively growing prompts alleviates catastrophic forgetting and increases model generalization capacities across tasks.

Many studies have explored how information storage functions within the Transformer architecture. \citet{elhage_mathematical_2022} gave a comprehensive overview of the Transformers architecture under the lens of mechanistic intepretability. \citet{geva_transformer_2021} showed that the feedforward layers of Transformers models act similarly to key-value memories in information retrieval systems. Based on that work, \citet{mitchell_fast_2021} introduced MEND, a framework that leverages a group of small networks to successfully perform local factual edits within the feedforward layers of a large Transformers model. \citet{meng_mass-editing_2022,meng_locating_2022} expanded on this idea by using causal inference to locate the attention feedforward layer containing a given fact and editing the corresponding matrix as a constrained optimization problem.

In contrast, several approaches for storing new information within a language model have been proposed. One such approach is the use of flexible external memories, as exemplified in \cite{wu_memorizing_2021,wu_efficient_2022}. Another, dynamic method is that of retrieval systems, which can leverage external knowledge bases, including the Web, to that purpose. Examples of such works include \cite{guu_realm_2020}, \cite{lewis_retrieval-augmented_2020}, \cite{borgeaud_improving_2021} and \cite{liu_reta-llm_2023}. Finally, new information can be stored in the short-term through methods such as prompt-tuning \citep{liu_gpt_2021,liu_p-tuning_2022}.

In terms of evaluation, \cite{petroni_language_2019} is an early attempt at measuring relational and factual knowledge within PLMs. \citet{zhu_modifying_2020} proposed new, information-theory based evaluation metrics for factual knowledge. \citet{kadavath_language_2022} and \citet{lin_teaching_2022} focused on measuring model uncertainty as a way to distinguish known facts from hallucinated ones. \citet{jang_towards_2021,jang_temporalwiki_2022} introduced the framework \textsc{TemporalWiki}, which like us, includes a process to generate datasets and benchmarks from information extracted from Wikipedia. However, their framework targets large scale continual learning while we focus on the factual knowledge acquisition point of view (detailed next). This difference in perspective leads to important differences in terms of types of inputs (facts vs texts), number of inputs, type and learning efficiency of the tested adaptation methods with respect to the number of parameters, and evaluation metrics (perplexity vs. factual MCQs accuracy). \citet{yu_kola_2023} detailed the creation of a large and refined benchmark, specifically tailored to measure world knowledge within PLMs. \citet{kasai_realtime_2022} proposed a continual MCQ benchmark for world knowledge, updated every week with new questions about recent events extracted from news websites.
\citet{yang_robust_2021} successfully used prefix-tuning to adapt a PLM for text classification, while \citet{ma_cpt_2022} used the same method for speech-to-text translation. Prefix-tuning was also shown to obtain good performance in natural language understanding \citep{lester_power_2021}, summarization \citep{chen_incorporating_2023} and sentiment analysis \citep{balakrishnan_exploring_2022} \textit{inter alia}. \citet{zhao_domain-oriented_2022} showed that prefix-tuning may also be used for efficient domain adaptation.

Parameter-efficient training methods, such as LoRa and prefix-tuning, are often used both 
to continue pretrain an LLM and to adapt it to a domain. However, 
recent works suggest that, with LoRa and full finetuning, very few new factual knowledge are actually learned~\citep{onebitlora}.
We propose in this work to investigate this question with prefix-tuning, which is based on similar principles than
in-context learning, a method that is known to be able to inject new knowledge.
Compared to the past litterature on prefix tuning, we focus on its properties with regard to
factual knowledge learning,
and give concrete answers to the questions of whether and when does prefix tuning learn new factual knowledge.

\section{Methodology}
\subsection{Research questions}
As shown in the related works Section, there is still not a clear understanding about what is really learnt by finetuning methods like LoRa.
In this study, we argue that prefix tuning is a better solution to inject a small number of new facts into the LLM, which may potentially be extended (in a future work) to support many facts either by retrieving 
the best prefix from a prefix-store (\`a la RAG), or by selecting prefixes with gating networks (\`a la mixture-of-experts)
or by generating prefixes with a dedicated model.
Concretely, the target research questions of this work are:
    (i) Can a single prefix vector on the first layer learn a single fact? Does this learning generalize to reformulations of this fact?
    (ii) Can a longer prefix ($n>1$) learn multiple facts? What effect does prefix size have on learning and generalization? In-context learning suggests that the answer to this question and the previous one are positive.
    (iii) In the existing literature, the prefix is usually spread across all layers of the model. However, \citet{simoulin_how_2021} suggest that the deeper layers in Transformer models are associated with abstract and high-level capabilities, while factual information is stored in the lower layers. Does restricting the prefix depth $d$ therefore affect the learning and generalization capacities of the model?
    (iv) Do the answers to the previous questions remain true with bigger models?

\subsection{Facts learning}
We model a fact as a semantic triple of the form (subject, predicate, object), in which the subject and object are typically noun phrases, and the predicate a verb phrase. 
We consider the following important properties, largely
adapted from \cite{meng_locating_2022}:\\
    {\bf Learning:} The updated LLM has learnt the fact when it can predict the object from a sentence containing the subject and predicate after being updated, while it could not predict the object before;\\
    {\bf Generalization:} The LLM is able to generalize the learned fact when it can predict the object from a paraphrase of the subject and predicate.\\
    {\bf Specificity:} The updated LLM is specific when it correctly generates another expected object that is different from the learned triplet from a slightly different subject and predicate input.\\
    {\bf Non-forgetting:} The updated LLM generates the correct objects that were already known by the baseline LLM.

\subsection{Evaluation}
Let $L$ be a baseline LLM and 
$T = T_1, ..., T_p$ a list of recent facts (triples).
We first build a training set 
containing a list of simple sentences generated from the triples in $T$ (see Figure~1).
We then update the model on this training set, either with prefix-tuning (our proposal) or LoRA (the baseline).
The perplexity of the updated LLMs are computed on the same training set and compared:
although it is largely debated in the community, we nevertheless consider that this perplexity is a relevant
indicator of whether the LLM has learnt this training set or not.
We then evaluate 
generalization by measuring the perplexity of the updated LLMs 
on complex, creative sentences created by reformulating the training sentences.
We finally 
measure specificity and non-forgetting by evaluating the LLMs on existing MCQ benchmarks.
    
\section{Dataset}
In this section, we describe the steps used to create \textsc{Novel-WD} and give an overview of the resulting dataset. A sample output of each step of the full process is given in Table \ref{tab:dataset_gen}.

\begin{table}[ht]
    \begin{center}
    {\scriptsize{
    \begin{tabular}{c|l}
         Element & Value\\
         \hline
         Triple & (Frances Allen, spouse, Jacob Schwartz)\\
         Training sentence & Frances Allen is married to Jacob Schwartz.\\
         Test sentence 1 & Frances Allen's spouse is\\
         Test sentence 2 & The spouse of Frances Allen was\\
         Test sentence 3 & Frances Allen was married to\\
         Test sentence 4 & Frances Allen has been married to\\
         Test sentence 5 & The name of Frances Allen's spouse is\\
         Question & Who was Frances Allen's spouse?\\
         Distractor 1 & Charles Householder\\
         Distractor 2 & David Padua\\
         Distractor 3 & John Cocke\\
    \end{tabular}
    }}
    \caption{A sample of the dataset for a single triple.}
    \label{tab:dataset_gen}
    \end{center}
\end{table}

\paragraph{Triple extraction}
We begin by extracting RDF triples that were newly added to Wikidata. To do so, we retrieve new triples from a daily incremental database dump. We restrict ourselves to items and exclude lexemes, which represent lexicographical data. We also do not take into account complex triples, in which the subject or object is a Wikimedia template, as well as triples in which the subject is a numerical identifier, a filename or a URI. We then resolve eventual internal Wikidata links in the subject, predicate or object by replacing them with the English name of the associated item. Finally, when multiple triples share the same subject and predicate, we randomly select one such triple and discard the other ones, so as to limit the risk of models trying to learn multiple conflicting facts.

\paragraph{Training set}
To generate a training set, we convert each triple into a simple sentence, by querying a 8-bit quantized version of \textsc{Vicuna-13b} \citep{chiang_vicuna_2023} with a two-shots prompt. For each triple, we generate one such sentence.

\paragraph{Two evaluation tasks}
The first evaluation is a causal language modeling task (perplexity): for each triple, we ask 8-bit \textsc{Vicuna-13b} in a two-shots setting to generate 5 sentences in which the object of the triple is missing. In order to test for generalization capabilities and to avoid repeating the training sentence, we specifically prompt {\it Vicuna} for "creative sentences". Manual editing may then be applied to the output sentences in the infrequent situation (occurring for less than 10 facts) where full sentences are generated rather than incomplete one.

The second task is a multiple choice question answering task (MCQ). For each triple, a two-shots 8-bit \textsc{Vicuna-13b} prompt is first applied to generate a question asking for the object of the triple. Then, a similar prompt is applied to generate 4 "likely answers" to the question. Among the 4 generated answers, we remove the ground-truth one if it is present, and select the 3 first remaining ones as distractors. After manually checking and editing the generated answers in rare cases (3 occurrences) where they semantically overlap, we then add in the correct answer. We therefore obtain a question with 4 possible choices, exactly one of which being correct.

After all the steps above have been applied, \textsc{Novel-WD} consists of 338 distinct triples, and each triple contains one associated training sentence, five incomplete validation sentences, one question and three distractors.
    
\section{Experimental setup}
The baseline model chosen for our experiments is \textsc{BLOOMZ-7.1b} \citep{muennighoff_crosslingual_2023}.
\textsc{BLOOMZ-7.1b} is a relatively old LLM, but which was particularly well designed:
all the fundamental architectural choices that equip recent LLMs were already there, including a large vocabulary size
that has also been adopted for instance in Gemma2. The few differences, such as grouped query attention, 
are designed to improve speed not performance,
so it is reasonable to assume that the behaviour observed for {\it BLOOMZ-7.1b} translates to similar LLMs.
Recent studies have also shown that, when appropriately finetuned, its performances matches those of state-of-the-art
LLMs~\cite{microsoftbloom}.
Its main drawback is it's small training data, but this is largely compensated by the fact that, in our view,
{\it all} of its data is known, which is a major advantage when aiming at rigorous scientific research.

The training was ran for up to 450 epochs using the AdamW optimizer with a weight decay of 0.1 and an initial learning rate of $3*10^{-2}$, decreasing by a factor of 10 after 10 epochs of non-decreasing training loss. We did not project the prefix through an intermediate MLP as mentioned in \cite{li_prefix-tuning_2021}, as we found that it did not increase training stability and generally resulted in lower performance.
For all of our models, prefix-tuning was implemented by learning the value of the previous key and value vectors in attention layers, resulting in two vectors per layer and per virtual token being learned, for a total of $2*d*n$ vectors.\\

For each macro-experiment and number of facts $k$, we divided the $D$=338 facts of \textsc{Novel-WD} into non-fully overlapping subsets of length $k$, and trained one copy of the baseline model on each subset. For a given $k$, the number of subsets was computed as $\max(5, \left \lfloor{D/k}\right \rfloor)$. For example, for $k=3$, we sampled 112 subsets of 3 facts, and trained a separate copy of \textsc{BLOOMZ-7b1} on each of those 112 subsets. Training subsets were generated for values of $k$ in $\{1, 2, 3, 4, 5, 8, 10, 20, 50, 100, 200\}$.

\subsection{Evaluation}
To evaluate our models in the text prediction setting, we prompt them with each of the five incomplete sentences associated with each fact from the training set, and generate the following ten tokens without sampling and with a temperature of 1. We only count an answer as correct if the model's output contains the exact answer's text, capitalization excepted, and we report the accuracy over every sentence of the test set for a given model. We also measure the {\bf{{proportion of learning models}}} for a given $k$, by selecting only facts of the test set for which the baseline model does not output any correct prediction, and counting the proportions of the prefix-tuned models trained on those questions for which the test set accuracy is non-zero. In other words, learning models are models which are able to correctly predict at least one sentence completion for facts that were not known by the baseline.

To perform regression tests, we selected the SciQ \citep{welbl_crowdsourcing_2017} and MMLU \citep{hendrycks_aligning_2020,hendrycks_measuring_2020} datasets. For SciQ, we measure the accuracy of the baseline and prefix-tuned models in the MCQ setting, by using the same prompt as for \textsc{Novel-WD}, and selecting the lowest per-token perplexity choice. We apply this method on all 1,000 questions of the test set. For MMLU, we append each of the possible four completions to each sentence, and then select the one with the lowest per-token perplexity as the model's answer. This is applied to the test sets from each of the 57 categories found in the dataset. Due to computational costs, regression tests were ran on a random sample of 5 prefix-tuned models for each value of $k$.
    
\section{Results and analysis}

\subsection{Base setup}
Our initial experiment focuses on a single prefix ($n=1, d=1$), corresponding to 8,192 trainable parameters, or 0.000116\% of the baseline model's parameters. For comparison, we also perform the same experiment using LoRA (rank$=8, \alpha=8$) instead of prefix-tuning. We use the same training hyperparameters for both LoRA and prefix-tuning.

The proportion of prefix-tuned models with increased accuracy in the prediction setting is given in Figure \ref{fig:acc_7b111}, along with the mean accuracy (see Appendix~B Figure~\ref{fig:acc_7b111b}) obtained in the prediction setting for different numbers of facts.

\begin{figure}
    \centering
    \includegraphics[width=\linewidth]{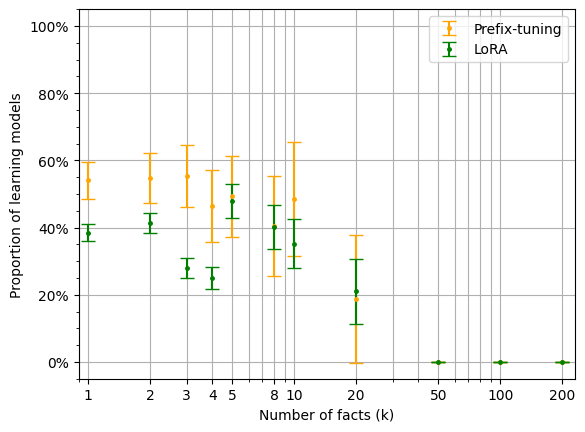}
    \caption{Percentage of prefix-tuned models obtaining increased accuracy over the baseline. Error bars span 95\% confidence intervals.}
    \label{fig:acc_7b111}
\end{figure}

For $1 \le k \le 3$, between 54.1\% and 55.4\% of the models are able to learn at least one information over the baseline. This amount stays stable for $k \le 10$, with the proportion of learning models ranging from 40.5\% to 55.4\%. For $k=20$, this proportion drops to 18.8\%, and none of the models trained for $k>20$ achieved any accuracy gains over the baseline.
Note that a recent work applying control theory to LLMs has shown that WikiText can be nearly 
perfectly predicted (at 97\%) with less than 10 additional prompt tokens~\cite{prompts10}, which also somehow confirms from a different
point of view this limit of $k\le 10$ tokens than we have found.

The baseline model obtains a consistent accuracy ranging from 3.0\% to 6.3\%, suggesting that a small number of facts found in the dataset are either already known or easily deducible by the model. In contrast, the prefix-tuned models obtain a mean accuracy peaking at 29.1\% for $k=3$, and gradually decreasing for $k>3$ until $k=50$, for which the results are no longer significantly better than the baseline.
This initial result suggests that during training, the prefix is usually able to select and remember 1 to 3 facts well, and up to 20 with decreasing accuracy. Furthermore, this learning is conditional on having a low enough number of facts present in the training data; having more than 10 facts seems to hamper the model's ability to learn even a single fact.

In comparison, models trained with LoRA systematically underperform prefix-tuned ones for all values of $k$, with a prediction accuracy reaching 20.4\% for $k=2$, and values ranging from 4.6\% to 14.4\% for other values of $k$. Furthermore, they typically obtain pLM scores that are similar or lower than the ones of prefix-tuned models. This may be due to the low rank value of 8 used in our experiments; however, rank 8 LoRA adds 3,932,160 parameters to the base model, a number which is 480 times higher than the parameters contained in a single prefix. We therefore argue that while LoRA may outperform prefix-tuning at higher matrix ranks, it does so in a much less cost-efficient manner than prefix-tuning.

\subsubsection{Error analysis}

With $k=1$, about half of the facts found in \textsc{Novel-WD} were not learned by a single prefix. While we could not identify meaningful semantic or content differences between the types of facts that were learned and those that were not, we report in Table \ref{tab:not_learned} in appendix~A quantitative statistics between those two categories. For each reported statistic, the non-learned value was found to be significantly larger than the learned one, as measured using a one-sided Welch's t-test (p = 0.05).
This suggests that the facts that were not successfully learned are typically longer and are farther from the baseline model's distribution, both in their sentence form and in the text completion setting, which might result in an inability for prefix-tuning to sufficiently steer the model towards learning them.

\subsection{Detecting overfitting and forgetting}

We report the training loss in Figure \ref{fig:7b111_loss_norm} and norm of the two prefix vectors in Appendix~B Figure~\ref{fig:7b111_loss_normb} measured post-training in each experiment.

\begin{figure}
    \centering
        \includegraphics[width=\linewidth]{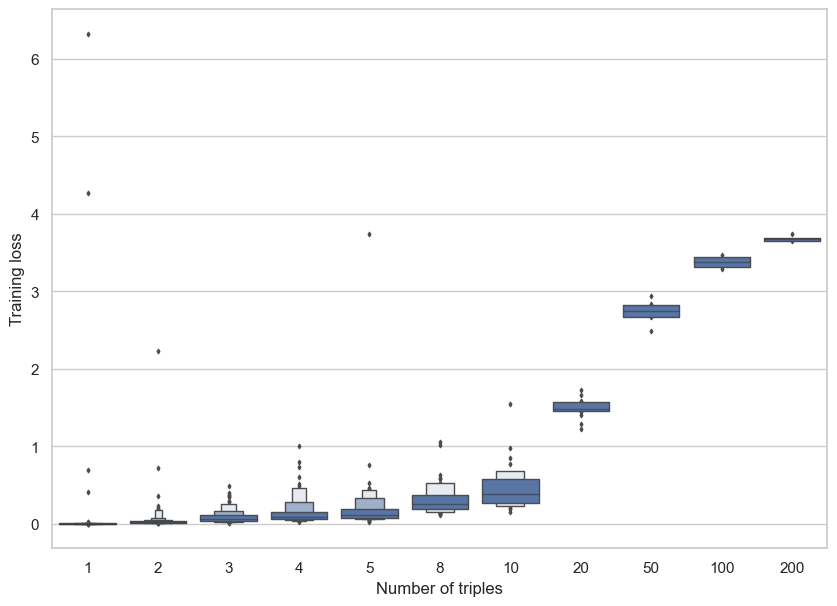}
    \caption{Training loss in the basic setup, measured post-training.}
    \label{fig:7b111_loss_norm}
\end{figure}

We observe that for $k=1$, almost all experiments end with a training loss approaching zero, with the exceptions of a few outliers for which the loss remains high. This confirms our previous finding that the prefix is almost always able to learn a single fact, but may not be able to generalize in the prediction setting. When increasing $k$, the losses increase linearly up to $k=10$ (median value: $L_{train} = 0.38$). For $n \ge 20$, the loss increases sharply and quickly approaches the baseline model's loss of 4.38. We interpret this inflection as consistent with our previous observations, suggesting that a change of learning mode occurs in the vicinity of $k = 15$: For lower values, the model is efficiently able to learn and generalize novel information, while for higher values, the model may no longer able to store all facts and instead unsuccessfully attempt to learn a combined representation of the training set. These findings are also consistent with the evolution of the prefix norm given: For $n \le 3$, we observe a linear increase in prefix norm, which may indicate that the model does not make full use of the available prefix capacity. For $3 \le n \le 10$, the prefix norm is nearly constant and may signal increasing compression within the prefix. Finally, for $n \ge 10$, the prefix norm decreases rapidly.

\begin{table}[htbp]
    \centering
    {\small{
    \begin{tabular}{c|c|c|c|c}
        & SciQ acc. & \multicolumn{3}{c}{MMLU acc.} \\
        k & & Min & Max & Avg \\
        \hline
        Baseline & 0.757 & 0.130 & 0.463 & 0.307 \\
        1        & 0.833 & 0.184 & 0.512 & 0.343 \\
        2        & 0.864 & 0.189 & 0.517 & 0.341 \\
        3        & 0.840 & 0.189 & 0.517 & 0.340 \\
        4        & 0.838 & 0.184 & 0.517 & 0.339 \\
        5        & 0.827 & 0.191 & 0.509 & 0.339 \\
        8        & 0.833 & 0.184 & 0.509 & 0.341 \\
        10       & 0.834 & 0.193 & 0.509 & 0.341 \\
        20       & 0.808 & 0.185 & 0.515 & 0.328 \\
        50       & 0.835 & 0.190 & 0.518 & 0.335 \\
        100      & 0.826 & 0.192 & 0.512 & 0.340 \\
        200      & 0.828 & 0.189 & 0.524 & 0.342 \\
    \end{tabular}
    }}
    \caption{Accuracy of the models on the MMLU and SciQ datasets, averaged over 5 random runs for each value of $k$. For MMLU, we report the score obtained by the lowest and highest accuracy as well as the average across categories.}
    \label{tab:regression}
\end{table}

Finally, we report in Table \ref{tab:regression} the results of the evaluation over SciQ and MMLU, which shows that the prefix-tuned models do not seem to forget facts learned during pre-training or incur any loss of reasoning capabilities, for any value of $k$. Surprisingly, our prefix-tuned models even perform consistently and significantly better than the baseline for all values of $k$. 
Our hypothesis is that, by "finetuning" (through a prefix) the LLM on Wikipedia-like sentences, we specialize the LLM to interpret its inputs in a more "factual way" and in the Wikipedia domain, which is useful for the type of factual MCQ questions that occur in SciQ and MMLU.
However, we did not study this hypothesis in detail and leave this question open for future work.


\subsection{Impact of prefix size}
Table \ref{tab:prefix_length} contains the results obtained when prefix-tuning \textsc{BLOOMZ-7b1} while varying the number of virtual tokens $n$ contained in the prefix.

\begin{table}[htbp]
    \centering
    {\small{
    \begin{tabular}{c|c|c|c|c|c|c}
        & \multicolumn{2}{c|}{n=1} & \multicolumn{2}{c|}{n=20} & \multicolumn{2}{c}{n=100} \\
        k & Acc & pLM & Acc & pLM & Acc & pLM \\
        \hline
        1   & 0.274 & 0.541 & \textbf{0.353} & \textbf{0.601} & \textbf{0.365} & \textbf{0.619} \\
        2   & 0.279 & 0.548 & \textbf{0.333} & 0.613          & \textbf{0.357} & 0.607 \\
        3   & 0.291 & 0.554 & 0.315          & 0.589          & \textbf{0.358} & 0.616 \\
        4   & 0.247 & 0.464 & \textbf{0.321} & \textbf{0.607} & \textbf{0.337} & \textbf{0.619} \\
        5   & 0.227 & 0.493 & \textbf{0.316} & 0.582          & \textbf{0.304} & 0.612 \\
        8   & 0.177 & 0.405 & \textbf{0.256} & 0.524          & \textbf{0.270} & 0.452 \\
        10  & 0.159 & 0.485 & \textbf{0.245} & 0.601          & \textbf{0.268} & 0.512 \\
        20  & 0.123 & 0.188 & \textbf{0.199} & \textbf{0.500} & \textbf{0.218} & \textbf{0.500} \\
        50  & 0.076 & 0     & \textbf{0.116} & 0.167          & \textbf{0.113} & 0.167 \\
        100 & 0.053 & 0     & \textbf{0.086} & \textbf{0.400} & \textbf{0.096} & \textbf{0.400} \\
        200 & 0.055 & 0     & 0.063          & 0              & 0.070         & 0 \\
    \end{tabular}
    }}
    \caption{Proportion of learning models (pLM) and mean prediction accuracy for different number of virtual tokens $n$ in the prefix. Bold values denote statistically significant improvements over $n=1$, using a one-sided z-test for proportions for pLM and a one-sided t-test for the accuracy $(p=0.05)$.}
    \label{tab:prefix_length}
\end{table}

We observe significant improvement in accuracy for nearly all values of $k$ when increasing the prefix size from 1 to 20, as well as significant gains in the proportion of learning models for $k \in \{1, 4, 20, 100\}$. Similar results are obtained when further increasing the prefix size from 1 to 100. However, none of the variation in accuracy or proportion of learning models between $n=20$ and $n=100$ are statistically significant.

We interpret those results as follows: Increasing the prefix size only modestly increases the chances for a model to be able to learn at least one fact. However, such an increase has a strong impact on the prediction capabilities of the model, which suggests that the model is able to learn more facts and to generalize better.

We hypothesize that the former may stem from the varying complexity of the facts in our dataset: for some facts, the base model may already contain information about the subject and predicate, and prefix-tuning might only be needed to learn the value of the object. A typical example of this situation can be found in facts of the type "[historical figure] was born on [date]". On the contrary, there exist more complex facts for which the subject and predicate themselves might be novel, and for which the base model might not contain information. We also note that increasing the prefix size past $20$ brings no further improvement to the learning and generalization capacities of our model, which may indicate that prefixes are inherently limited in terms of information capacity.

\subsection{Impact of prefix depth}

We report in Table \ref{tab:prefix_depth} the results obtained by increasing the number of layers spanned by the prefix in our initial setup from $d=1$ (minimal depth) to $d=30$ (full-depth prefix).

\begin{table}[htbp]
    \centering
    {\small{
    \begin{tabular}{c|c|c|c|c}
        & \multicolumn{2}{c|}{d=1} & \multicolumn{2}{c}{d=30} \\
        k & Acc & pLM & Acc & pLM \\
        \hline
        1   & 0.274 & 0.541 & \textbf{0.354} & 0.590 \\
        2   & 0.279 & 0.548 & \textbf{0.441} & \textbf{0.667} \\
        3   & 0.291 & 0.554 & \textbf{0.520} & \textbf{0.768} \\
        4   & 0.247 & 0.464 & \textbf{0.467} & \textbf{0.690} \\
        5   & 0.227 & 0.493 & \textbf{0.470} & \textbf{0.731} \\
        8   & 0.177 & 0.405 & \textbf{0.487} & \textbf{0.690} \\
        10  & 0.159 & 0.485 & \textbf{0.476} & \textbf{0.789} \\
        20  & 0.123 & 0.188 & \textbf{0.401} & \textbf{0.813} \\
        50  & 0.076 & 0     & \textbf{0.275} & \textbf{0.333} \\
        100 & 0.053 & 0     & \textbf{0.130} & \textbf{0.800} \\
        200 & 0.055 & 0     & \textbf{0.101} & \textbf{0.000} \\
    \end{tabular}
    }}
    \caption{Proportion of learning models (pLM) and mean prediction accuracy for different prefix depths $d$ in the prefix. Bold values denote statistically significant improvements over $d=1$, using a one-sided z-test for proportions for pLM and a one-sided t-test for the accuracy $(p=0.05)$.}
    \label{tab:prefix_depth}
\end{table}

We observe that increasing the prefix depth has a significant effect on both the accuracy and the proportion of learning models. For all values of $k$, the average accuracy is increased by 8 to 31\%, with the highest increase reached for $k=10$. The highest average accuracy is obtained for $k=3$, which once more suggests that up to three facts can be efficiently stored within a prefix, but performance stays comparable up to $k=10$.

The second main observation is the fact that the proportion of learning models significantly increases for all values of $k$ except $k=1$, with gains of up to 80\% for $k=100$. We hypothesize that increasing the prefix depth allows for more complex information to be learned and enables the model to learn at least one information for all but the highest amount of training facts.
Increasing the value of $d$ from $1$ to $30$ effectively multiplies the number of trainable parameters by 30, but far surpasses the results obtained by increasing the prefix length by a factor of 100. We therefore remark that prefix depth seems to have a much stronger effect on model performance than prefix length.

\subsection{Impact of base model}

To investigate the effect that the type and size of the base model may have on prefix-tuning, we repeat our initial experiments on \textsc{BLOOMZ-1b7}, the 1.7 billion parameter version of BLOOMZ, chosen for scale comparisons.
We measure the accuracy of the baseline models in the prediction setting over the entirety of \textsc{Novel-WD}. \textsc{BLOOMZ-1b7} obtained an overall accuracy of 4.4\%, while \textsc{BLOOMZ-7b1} reached a similarly low value of 5.0\%.

The results obtained after prefix-tuning are reported in Table \ref{tab:base_model}.
\begin{table}[htbp]
    \centering
    {\small{
    \begin{tabular}{c|c|c|c|c}
        & \multicolumn{2}{c|}{\textsc{BLOOMZ-1b7}} & \multicolumn{2}{c}{\textsc{BLOOMZ-7b1}} \\
        k & Acc & pLM & Acc & pLM \\
        \hline
        1   & 0.293 & 0.565 & 0.274          & 0.541 \\
        2   & 0.273 & 0.556 & 0.279          & 0.548 \\
        3   & 0.262 & 0.589 & 0.291          & 0.554 \\
        4   & 0.213 & 0.464 & \textbf{0.247} & 0.464 \\
        5   & 0.189 & 0.403 & \textbf{0.227} & 0.493 \\
        8   & 0.152 & 0.286 & 0.177          & 0.405 \\
        10  & 0.112 & 0.394 & \textbf{0.159} & 0.485 \\
        20  & 0.085 & 0.189 & \textbf{0.123} & 0.188 \\
        50  & 0.053 & 0     & 0.076          & 0     \\
        100 & 0.045 & 0     & 0.053          & 0     \\
        200 & 0.039 & 0     & \textbf{0.055} & 0     \\
    \end{tabular}
    }}
    \caption{Proportion of learning models (pLM) and mean prediction accuracy for different number of virtual tokens $n$ in the prefix. Bold values denote statistically significant improvements over the previous column, using a one-sided z-test for proportions for pLM and a one-sided t-test for the accuracy $(p=0.05)$.}
    \label{tab:base_model}
\end{table}
In terms of scaling, we first note that there are no significant improvements in terms of the proportion of learning models between \textsc{BLOOMZ-1b7} and \textsc{BLOOMZ-7b1}. This strengthens the intuition that this may be due to the inherent complexity of some facts in the dataset, and to the fact that the ability to learn a fact is already present in smaller models. However, increasing the model size has a noticeable effect on the prediction accuracy, which increases by several percentage points for $k \in \{4, 5, 10, 20, 50\}$. We believe that this is partially due to the scaling generalization capabilities of the models. However, as the number of trainable parameters almost doubles between \textsc{BLOOMZ-1b7} and \textsc{BLOOMZ-7b1}, these improvements may also be explained by an increase in prefix capacity.

Finally, to give an idea of the extracted facts, the quality of the synthetic generated sentences and which facts are correctly classified by the baseline model, 
Table~\ref{tab:ex} in Appendix~C shows a random extract of known facts and generated sentences:
some facts may "leak" from the LLM pretraining corpus (e.g., {\it Frederik Storm in Denmark}),
or may be guessed (e.g., {\it Vitale Faliero, language spoken, Italian})
or may be answered by chance (e.g., {\it A View to a Kill, MPA rating, PG}).
This question of leakage vs. actual forecasting is discussed in more details in~\cite{forecasting}.

\section{Conclusion}
In this study, we have developed a dataset for novel fact learning in pre-trained language models. We have shown that prefix-tuning can be used to learn new facts, and investigated the effect of various factors on prefix-tuning performance. Our main recommendation is to use full-depth prefixes, but to limit the prefix length to 20 virtual tokens.

We see several major avenues for future research based on this work. While we measured the effect of different factors independently, their combined effect might be different. In particular, it is hard to predict how prefix length and depth may interact together. Another research direction is the use of different and more recent baseline architectures such as Mixtral \citep{jiang2024mixtral}. Finally, a long-term goal could be to scale our approach to larger datasets, for example by using a mixture of prefixes at capacity along with a routing module. This could allow the use of a small, regular stream of new information to continually update a model.

\section{Limitations}

While this paper addresses the challenge of updating LLMs with novel facts, there are other types of "updates" that
should be achieved to make the updated LLM as useful as a new LLM pretrained from scratch on an up-to-date corpus,
such as language and topic drifts. The method described in this work can not solve this issue.
More generally, representing knowledge with triples is very limited, and can hardly for instance encode
time-dependent and location-dependent cultural preferences, common sense and beliefs.
This work is thus strongly limited in terms of the type of knowledge it can capture, but it is only a first step
towards a more general LLM updating paradigm.

Another limitation is that only a few facts are injected in the LLM with our method, while continual updating of
the LLM would require a constantly increasing number of facts to be added. To achieve this, our method would require
an additional step to select or generate the appropriate prefixes, depending on the observed context, in a similar way
as what is done with RAG or alternatively mixture of experts. We have not tested in this work such an enhancement,
and we have only focused so far on studying the usefulness of prefix tuning as an alternative to RAG and LoRA.

Finally, an apparent limitation may be the size of \textsc{Novel-WD}, which is quite small. However, this is mainly
because of the high cost of running the large number of experiments required in this study.
However, since 2020, Wikidata grows at a rate of 7 million entities per year (see {\url{https://en.wikisource.org/wiki/Wikidata_The_Making_Of}}), and the filtering that we apply leads to about 32000 remaining new facts per day (as checked for 14th March 2024), so getting data at scale should not be an issue.
Furthermore, although we made a few manual interventions to check for generation errors when creating the dataset and benchmarks,
we are convinced such interventions could be avoided when using better LLM, such as Llama3-70b or Qwen-72b.

\bibliography{bibs}

\newpage

\appendix

\section{Learned and non-learned facts}

Table~\ref{tab:not_learned} gives some statistics about the facts that have been learned or not learned in our experiments.

\begin{table*}[htbp]
    \centering
    \begin{tabular}{l|c|c|c|c}
        & \multicolumn{2}{c|}{Train set facts} & \multicolumn{2}{c}{Test set facts} \\
        Metric & Non-learned & Learned & Non-learned & Learned\\
        \hline
        Length (characters) & 57.8 & 51.0 & 73.5 & 66.2 \\
        Length (tokens) & 15.5 & 13.3 & 18.2 & 15.9 \\
        Length of $o$ (characters) & 17.8 & 15.6 & - & - \\
        \textsc{BLOOMZ-7b1} per-token perplexity & 4.56 & 4.30 & 4.26 & 4.18 \\
    \end{tabular}
    \caption{Quantitative comparison of the facts of \textsc{Novel-WD} that were successfully learned and those which were not within a single prefix. Reported values are averaged per category.}
    \label{tab:not_learned}
\end{table*}

\section{Impact from the number of novel facts}

\begin{figure}[h!]
        \centering
        \includegraphics[width=\linewidth]{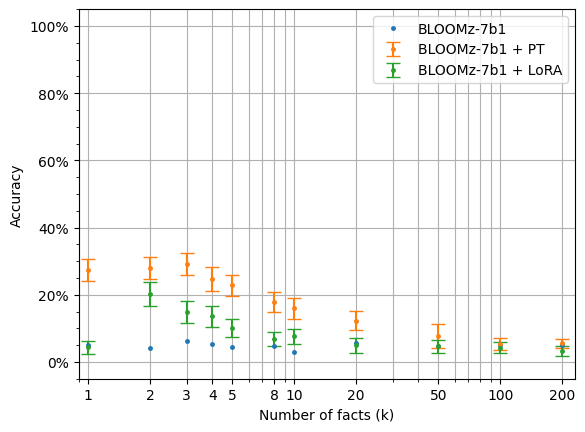}
    \caption{Mean accuracy of prefix-tuned (PT) models, LoRA models and of the baseline (right) in the prediction setting. Error bars span 95\% confidence intervals.}
    \label{fig:acc_7b111b}
\end{figure}

Figure~\ref{fig:acc_7b111b} complements Figure~\ref{fig:acc_7b111} by showing the mean accuracy of the models as a function of the number of facts, confirming the diminushing returns when increasing the number of new facts beyond 10.

\begin{figure}[h!]
        \centering
        \includegraphics[width=\linewidth]{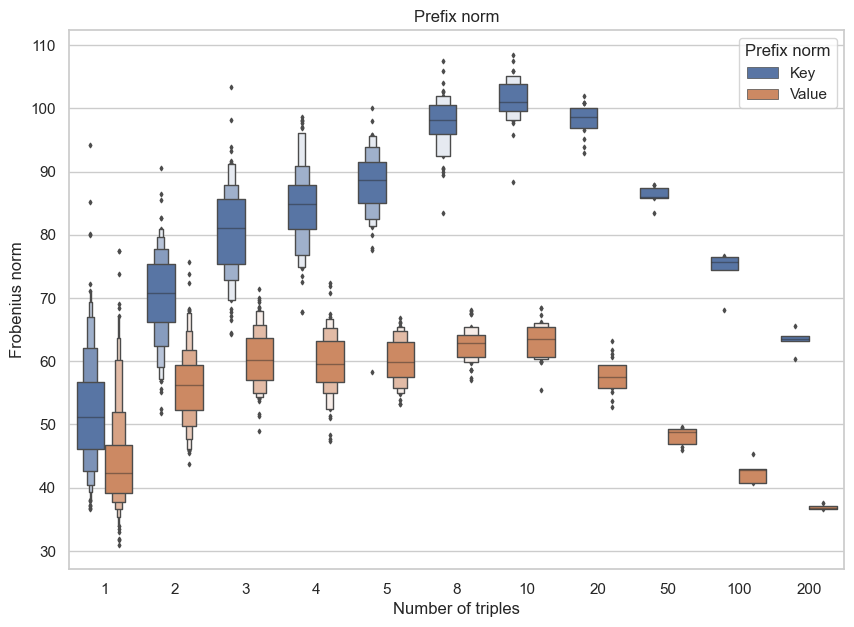}
    \caption{Frobenius norm of the key and value vectors of the prefix in the basic setup, measured post-training.}
    \label{fig:7b111_loss_normb}
\end{figure}

Figure~\ref{fig:7b111_loss_normb} complements Figure~\ref{fig:7b111_loss_norm} by showing the observation of two phases
with less and more than 10 new facts.

\section{Qualitative examples}

Table~\ref{tab:ex} shows both examples of generated sentences and facts that are already known by the model.
All of these samples have been randomly extracted, without any cherry picking.

\begin{table}[htbp]
    \centering
    {\small{
    \begin{tabular}{l}
        The Lesser hairy-footed dunnart is also known as S. youngsoni. \\
        Milady de Winter died by homicide. \\
        Garden Warbler is also known as S. borin. \\
        Dylan and Cole Sprouse were born on 4 August 1992. \\
        Yannick Aguemon is 180 centimetres tall. \\
        Heinrich Hoffmann died of natural causes. \\
\hline
 Chen Lin, occupation, writer \\
 White Flag, language of work or name, English \\
 A View to a Kill, MPA rating, PG \\
 Corey Hart, language spoken, English \\
 Extinction, mitigated by, conservation efforts \\
 Frederik Storm, country for sport, Denmark \\
    \end{tabular}
}}
    \caption{Random samples of generated sentences (top) and "already known" facts (bottom)}
    \label{tab:ex}
\end{table}

\end{document}